\documentclass[12pt,a4paper]{article}
\usepackage[dvipsnames, table]{xcolor}

\usepackage[T1]{fontenc}
\usepackage[utf8]{inputenc}
\usepackage{XCharter}

\usepackage{microtype}
\usepackage[margin=1in]{geometry}

\usepackage{graphicx}
\usepackage{subcaption}
\usepackage[normalem]{ulem}
\usepackage{enumitem}
\usepackage[numbers, sort&compress]{natbib}
\usepackage{dirtytalk}

\usepackage{physics}
\usepackage{nicefrac}

\definecolor{cvprblue}{rgb}{0.21,0.49,0.74}
\usepackage[pagebackref, breaklinks, colorlinks, allcolors=cvprblue]{hyperref}

\usepackage[capitalize]{cleveref}
\usepackage{pgfplots}
\pgfplotsset{compat=1.18}
\usetikzlibrary{external}
\AddToHook{env/algorithmic/begin}{\tikzexternaldisable}

\usepackage{mathtools}
\usepackage{amssymb}
\usepackage{bm}

\usepackage{algorithm}
\usepackage{algpseudocodex}

\usepackage{booktabs}
\usepackage{multirow}

\usepackage{authblk}

\date{\vspace{-1em}}

\title{Coverage-Constrained Human-AI Cooperation with Multiple Experts}

\author[1]{Zheng Zhang}
\author[1]{Cuong Nguyen}
\author[1]{Kevin Wells}
\author[2]{Toan Do}
\author[1,3]{David Rosewarne}
\author[1]{Gustavo Carneiro}
\affil[1]{Centre for Vision, Speech and Signal Processing, University of Surrey, United Kingdom}
\affil[2]{Department of Data Science and AI, Monash University, Australia}
\affil[3]{Royal Wolverhampton Hospitals NHS Trust, UK}
\begin{document}
\maketitle
\begin{abstract}
Human-AI cooperative classification (HAI-CC) approaches aim to develop hybrid intelligent systems that enhance decision-making in various high-stakes real-world scenarios by leveraging both human expertise and AI capabilities. Current HAI-CC methods primarily focus on learning-to-defer (L2D), where decisions are deferred to human experts, and learning-to-complement (L2C), where AI and human experts make predictions cooperatively. However, a notable research gap remains in effectively exploring both L2D and L2C under diverse expert knowledge to improve decision-making, particularly when constrained by the cooperation cost required to achieve a target probability for AI-only selection (i.e., coverage). In this paper, we address this research gap by proposing the \textbf{C}overage-constrained \textbf{L}earning to \textbf{D}efer and \textbf{C}omplement with Specific Experts (CL2DC) method. CL2DC makes final decisions through either AI prediction alone or by deferring to or complementing a specific expert, depending on the input data. Furthermore, we propose a coverage-constrained optimisation to control the cooperation cost, ensuring it approximates a target probability for AI-only selection. This approach enables an effective assessment of system performance within a specified budget. Also, CL2DC is designed to address scenarios where training sets contain multiple noisy-label annotations without any clean-label references. Comprehensive evaluations on both synthetic and real-world datasets demonstrate that CL2DC achieves superior performance compared to state-of-the-art HAI-CC methods.
\end{abstract}

\section{Introduction}
\label{sec:intro}
Machine learning models are becoming increasingly critical in real-world scenarios due to their high efficiency and accuracy. However, in high-stakes situations like risk assessment~\citep{green2019disparate}, content moderation~\citep{lykouris2024learning}, breast cancer classification~\citep{halling2020optimam}, and the detection of inaccurate or deceptive content produced by large language models~\citep{de2018hate,wei2022emergent,bubeck2023sparks,mozannar2024effective,ding2024hybrid}, human experts often provide more reliable and safer predictions compared to AI models.
To address the trade-off between human expertise and AI capabilities, \emph{human-AI cooperative classification} (HAI-CC) methods have been developed~\citep{punzi2024ai,dafoe2021-cooperative,steyvers2022bayesian}. These HAI-CC approaches enhance not only the accuracy, interpretability, and usability of AI models but also improve user efficiency and decision consistency over manual processes, significantly reducing human error~\citep{dafoe2021-cooperative,steyvers2022bayesian}.

HAI-CC approaches~\citep{dafoe2021-cooperative} aim to develop a \emph{hybrid intelligent} system that maximises accuracy while minimising the cooperation costs with \emph{learning-to-defer} (L2D) and \emph{learning-to-complement} (L2C) strategies. In L2D~\citep{mozannar2020consistent}, HAI-CC strategically decides when to classify with the AI model or defer to human experts, while L2C~\citep{complement_wilder} combines the predictions of AI and human experts.

When facing with challenging or high-stakes decisions, single-expert HAI-CC (SEHAI-CC) systems allow the system to defer to or complement with a fixed expert~\citep{narasimhan2022post,raghu2019algorithmic,okati2021differentiable,mozannar2020consistent,verma2022calibrated,whoshould_mozannar23,complement_wilder,charusaie2022sample,charusaiedefer,mao2024realizable,tariq2024a2c,charusaie2024unifying}. 
However, given the diverse range of expertise of different professionals, relying solely on a single expert for decisions across all input cases is impractical and potentially suboptimal. To address this, multiple-expert HAI-CC (MEHAI-CC) methods have been proposed to explore strategies for either complementing or deferring decisions to one or several experts simultaneously~\citep{ijcai2022-344,verma2022calibration,mao2024two,multil2d,zhang2023learning,lecodu,mao2024regression,tailor2024learning,keswani2021towards,alves2024cost}, effectively leveraging diverse expert knowledge for more robust decision-making.
Nevertheless, a remarkable gap in such MEHAI-CC approaches is that they rarely address L2D and L2C concomitantly, and even when they do consider both tasks in a single approach~\citep{lecodu}, they effectively disregard diverse expert knowledge by randomly selecting experts for the cooperative classification.

\begin{figure}[t]
    \centering
    \includegraphics[width=0.65\linewidth]{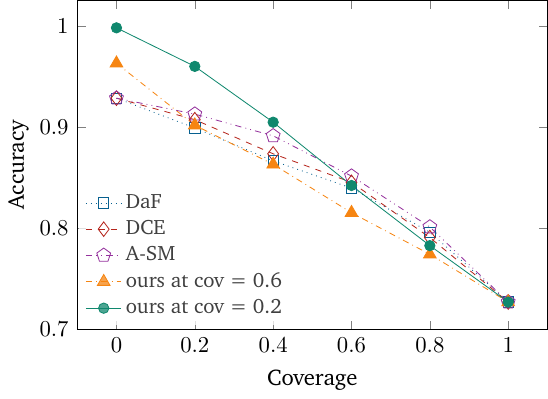}
    \caption{The post-hoc analysis to generate the coverage - accuracy curves of our proposed method on the Chaoyang dataset~\citep{zhu2021hard} is unreliable because the same method trained with different coverage constraints produces different curves. When comparing to several HAI-CC methods~\citep{charusaiedefer, dce, cao2024defense} plotted with the same post-hoc approach, it is possible to select the curve showing the best coverage - accuracy result, which may present an overly optimistic assessment of the method's performance. For instance, our method trained for two different coverages (i.e., 0.6 in \textcolor{BurntOrange}{orange} and 0.2 in \textcolor{PineGreen}{green}) show quite different performances.
    }
    \label{fig:posthoc_chaoyang}
\end{figure}

Another crucial issue in HAI-CC methods is the trade-off between accuracy and cooperation cost as it reflects the system's efficiency and effectiveness. Existing HAI-CC methods often analyse such trade-off through accuracy-coverage curves to evaluate performance at different coverage levels~\citep{narasimhan2022post,whoshould_mozannar23,lecodu}. Coverage is defined as the \textit{percentage of examples classified by the AI model alone}, where 100\% coverage indicates that all classifications are performed by the AI, and 0\% coverage means that all classifications are handled exclusively by experts.
HAI-CC methods~\citep{whoshould_mozannar23,narasimhan2022post} are typically trained with optimisation functions that  aim to balance accuracy and cooperation cost. However, in practice, the training process is brittle; small adjustments to the hyper-parameter controlling accuracy and cooperation cost often result in coverage values collapsing to either 0\% or 100\%~\citep{lecodu,zhang2023learning}.
Importantly, this hyper-parameter does not set a specific coverage value but rather allows for only a rough adjustment of cost influence within the optimisation, making it challenging to achieve a precise coverage target.
As a result, existing HAI-CC methods~\citep{whoshould_mozannar23,narasimhan2022post} often employ \emph{post-hoc} technique to construct accuracy - coverage curves by sorting deferral scores and adjusting the prediction threshold to obtain the accuracy at the expected coverage. 
This post-hoc approach is impractical, as it requires access to all testing samples before making predictions and lacks the ability to set and evaluate workload control during training. 
In addition, using the post-hoc approach to analyse the coverage - accuracy of a model trained in one coverage setting is unreliable. For example, \cref{fig:posthoc_chaoyang} illustrates that models of the same method, but trained under different levels of coverage constraints, yield different curves using the post-hoc method (e.g., the \textcolor{BurntOrange}{orange} and \textcolor{PineGreen}{green} curves). Hence, reporting the result by simply selecting the best performing method does not represent a reliable assessment of the approach.
Therefore, further research is needed to develop a principled mechanism for managing workload distribution in HAI-CC methods.
 
In this paper, we propose the novel \textbf{C}overage-constrained \textbf{L}earning to \textbf{D}efer and \textbf{C}omplement with Specific Experts (CL2DC) method. 
CL2DC integrates the strengths of L2D and L2C, particularly in training scenarios with multiple noisy-label annotations, enabling the system to either make final decisions autonomously or cooperate with a specific expert.
CL2DC not only determines when to defer to or complement with experts but also assesses the specific expertise of each expert, selecting the most suitable one for the decision-making process. We also introduce an innovative coverage constraint penalty in the loss function to effectively control coverage levels. This penalty enables a robust training process that reliably achieves the target coverage, allowing for a consistent and meaningful analysis of various methods using coverage-accuracy curves.
Our main contributions are summarised as follows:
\begin{itemize}
    \item We propose the CL2DC that integrates L2D and L2C strategies, enabling deferral to or complementation with specific experts in the presence of multiple noisy-label annotations.
    \item We introduce an innovative coverage constraint into our training process, targeting specific coverage values to effectively manage the  trade-off between coverage and accuracy in HAI-CC.
\end{itemize}

We evaluate our CL2DC method against state-of-the-art (SOTA) HAI-CC methods~\citep{multil2d,dce,charusaiedefer,cao2024defense,lecodu} using real-world and synthetic multi-rater noisy label benchmarks, such as CIFAR-100~\citep{wei2021learning,barz2020we}, Galaxy Zoo~\citep{bamford2009galaxy}, HAM10000~\citep{tschandl2018ham10000}, NIH-ChestXray~\citep{majkowska2020chest,wang2017chestx}, MiceBone~\citep{schmarje20192d,schmarje2022one,schmarje2022data}, and Chaoyang~\citep{zhu2021hard}. Results show that CL2DC consistently outperforms  previous HAI-CC methods with higher accuracy for equivalent coverage values for all benchmarks.

\section{Related work}
\label{sec:related}

\subsection{Human-AI Cooperative Classification}

HAI-CC approaches~\citep{dafoe2021-cooperative} seek to develop a \emph{hybrid intelligent} system to maximise the cooperative accuracy beyond what either AI models or human experts can achieve independently, while simultaneously minimising the cooperation costs through \emph{learning-to-defer} (L2D) and \emph{learning-to-complement} (L2C) strategies. 

\textbf{Learning to Defer (L2D)} is an extension of \emph{rejection learning}~\citep{cortes2016learning}, which aims to learn a classifier and a rejector to decide in which case the decision should be deferred to a human expert to make the final prediction~\citep{madras2018predict, keswani2021towards, narasimhan2022post, mao2023two}. Existing L2D approaches focus on the development of different surrogate loss functions to be consistent with the Bayes-optimal classifier~\citep{narasimhan2022post, charoenphakdee2021classification, raghu2019algorithmic, okati2021differentiable, mozannar2020consistent, verma2022calibrated, whoshould_mozannar23, charusaie2022sample, cao2024defense, straitouri2023improving, liumitigating, mozannar2022teaching,charusaie2024unifying}. \citet{dce} explore the dependence between AI and human experts and propose a dependent Bayes optimality formulation for the L2D problem. However, these methods overlook practical settings in which there is a wide diversity of multiple human experts. Given such an issue, recent research in L2D shifts towards the multiple-expert setting~\citep{verma2022calibration, mao2023two, multil2d, keswani2021towards, babbar2022utility, mao2023principled, hemmer2023learning, tailor2024learning, leitao2022human,mao2024regression,zhang2023learning,alves2024cost}.
For example, \citet{multil2d} proposed a L2D method to defer the decision to one of multiple experts. \citet{mao2024regression} addressed both instance-dependent and label-dependent costs and proposed a novel regression surrogate loss function. Nevertheless, current research in L2D does not consider options that aggregate the predictions of human experts and AI model to make a joint decision.

\textbf{Learning to Complement (L2C)} methods aim to optimise the cooperation between human experts and the AI model by combining their predictions~\citep{complement_wilder,steyvers2022bayesian,kerrigan2021combining,liu2023humans,bansal2021most,ijcai2022-344,keswani2021towards,tariq2024a2c,charusaiedefer,lecodu,zhang2023learning}. 
\citet{liu2023humans} leverage perceptual differences through post-hoc teaming, showing that human - machine collaboration can be more accurate than machine - machine collaboration.
Recently, \citet{charusaiedefer} introduce a method that determines whether the AI model or a human expert should predict independently, or if they should collaborate on a joint prediction -- this is effectively a combined L2D and L2C approach, but it is limited to single expert setting.
\citet{ijcai2022-344} introduce a model featuring an ensemble prediction involving both AI and human predictions, yet it does not optimise the cooperation cost. 
\citet{lecodu} propose a combine L2D-L2C approach that integrates AI predictions with multiple random human experts, but overlooking the expert specificity.

\subsection{Learning with Noisy Labels}
\label{sec:LNL_review}

The vast majority of HAI-CC methods assume that the ground-truth \emph{clean} annotations are available in the training set. The vast majority of HAI-CC methods assume that the ground-truth clean annotations are available in the training set. However, such assumption is not warranted in practice, particularly in applications like medical imaging, where a definitive \emph{clean} label may not be available due to the absence of final pathology. Therefore, we can only access expert opinions, which means multiple noisy annotations per training sample. Only recently, several HAI-CC systems have been designed to handle noisy-label problems. Here, we provide a short review of learning with noisy labels (LNL) methods that can be used in the HAI-CC context~\citep{song2022learning,carneiro_machine_2024}.

LNL approaches have explored many techniques, including: robust loss functions~\citep{zhang2018generalized,ghosh2017robust}, 
co-teaching~\citep{MentorNet, han2018co},
label cleaning \citep{yuan2018iterative, jaehwan2019photometric}, 
semi-supervised learning (SSL)~\citep{ li2020dividemix, ortego2021multi}, 
iterative label correction~\citep{label_clean,arazo2019unsupervised}, 
meta-learning ~\citep{L2W,Distill_noise,FSR,Famus}, diffusion models~\citep{chen2024label}, and 
graphical models~\citep{garg2023instance}.
Among the top-performing LNL methods, we have
ProMix~\citep{wang2022promix} that introduces an optimisation based on a matched high-confidence selection technique. Another notable LNL approach is DEFT~\citep{wei2024vision} that utilises the robust alignment of textual and visual features, pre-trained on millions of auxiliary image-text pairs, to sieve out noisy labels. 
Despite achieving remarkable results, LNL with a single noisy label per sample suffers from the identifiability issue~\citep{liu2023identifiability}, meaning that a robust training may require multiple noisy labels per samples.
Methods that can deal with  multiple noisy labels per sample are generally known as multi-rater learning (MRL) approaches.


MRL aims to train a robust classifier with multiple noisy labels from multiple human annotators, which can be divided into inter- and intra-annotator approaches. The inter-annotator methods focus on characterising the variabilities between annotators~\citep{raykar2009supervised, guan2018said, mirikharaji2021d, ji2021learning}, while the intra-annotator approaches focus on estimating the variability of each annotator~\citep{khetan2017learning, tanno2019learning, wu2022learning, cao2023learning}. 
Recently, UnionNet~\citep{wei2022deep} has been developed to integrate all labelling information from multiple annotators as a union and maximise the likelihood of this union through a parametric transition matrix. 
Annot-Mix~\citep{herde2024annot} extend mixup and estimates each annotator’s performance during the training process.
CROWDLAB~\citep{goh2022CROWDLAB} is a state-of-the-art (SOTA) MRL method that produces consensus labels using a combination of multiple noisy labels and the predictions by an external classifier. 

\section{Methodology}
\label{sec:method}
    \begin{figure*}[t]
    \centering
    \includegraphics[width=\linewidth]{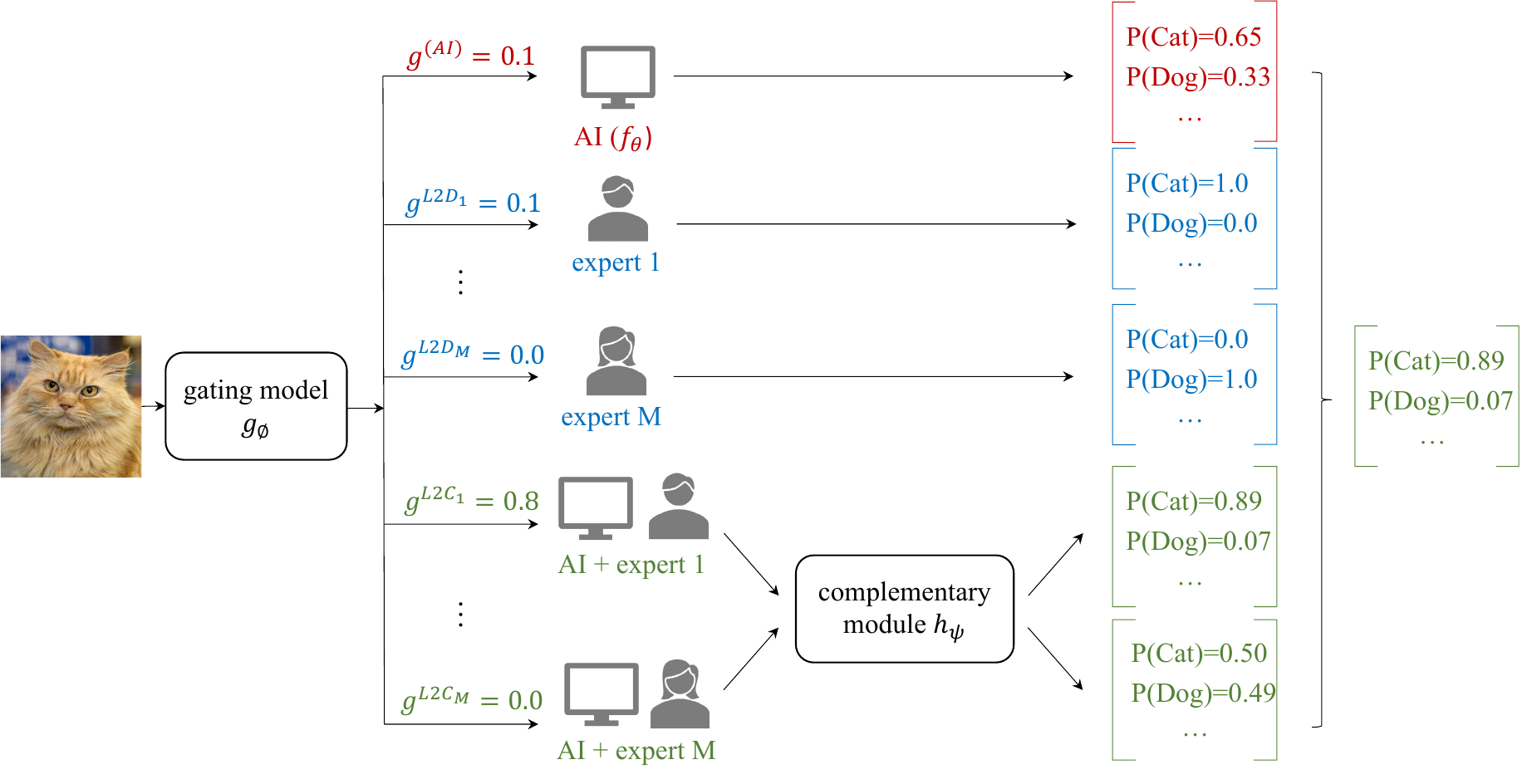}
    \caption{CL2DC contains a gating model \(g_{\phi}(.)\), a complementary module \(h_{\psi}(.)\), and an AI model \(f_{\theta}(.)\). The gating model aims to decide whether we use LNL-trained AI model \(f_{\theta}(.)\) alone (i.e., when \(g^{(\mathrm{AI})}_{\phi}(.)\) has the largest probability), defer the decision to one of the \(M\) experts $\{1,\dots,M\}$ (i.e., when one of the \(g^{(\mathrm{L2D}_{j})}_{\phi}(.)|_{j=1}^{M}\) has the largest probability), or complement the LNL AI model's prediction, through the complementary module,  with one of the \(M\) experts  (i.e., when one of \(g^{(\mathrm{L2C}_{j})}_{\phi}(.)|_{j=1}^{M}\) has the largest probability). In the figure, the gating model selects L2C between AI and user 1,  given its largest probability of \(0.8\), to make the final prediction, on the right.}
    \label{fig:architecture}
\end{figure*}
    Let $\mathcal{D}=\{\mathbf{x}_i,\mathcal{M}_i\}^{N}_{i=1}$ be the noisy-label multi-rater training set of size \(N\), where $\mathbf{x}_i \in \mathcal{X} \subset \mathbb{R}^{H \times W \times R}$ denotes an input image of size $H\times W$ with $R$ channels, and $\mathcal{M}_i=\{ m_{i,j}: m_{i, j} \in \mathcal{Y}= \{1, \dots, C\} \}_{j=1}^{M}$ denotes the noisy annotations of \(M\) human experts for the input image $\mathbf{x}_{i}$, with \(C\) being the number of classes. 
    
    Our proposed method contains three components, an AI classifier, a gating model, and a complementary module,  which form an adaptive decision system that leverages AI strengths while maintaining human oversight, ensuring a balance between performance and trustworthiness in complex decision environments.
    More specifically, we have: 1) an \emph{AI classifier} $f_\theta:\mathcal{X} \to \Delta^{C-1}$, parametrised by \(\theta \in \Theta\), where \(\Delta^{K - 1} = \{\mathbf{p}: \mathbf{p} \in [0, 1]^{K} \wedge \pmb{1}^{\top} \mathbf{p} = 1\}\) denotes the \((K - 1)\)-dimensional probability simplex; 2) a \emph{gating model} \(g_{\phi}: \mathcal{X} \to \Delta^{2M + 1}\), parameterised by \(\phi \in \Phi\), which produces a categorical distribution reflecting the probability of selecting the prediction made by the AI model alone (i.e., \(g^{(\mathrm{AI})}_{\phi}(.)\)), or deferring the decision to one of the \(M\) human experts (i.e., \(g^{(\mathrm{L2D}_{j})}_{\phi}(.)|_{j=1}^{M}\) denote the probability of selecting expert 1 through expert $M$), or performing a complementary classification between the AI model and one human expert (i.e., \(g^{(\mathrm{L2C}_{j})}_{\phi}(.)|_{j=1}^{M}\) represent the probability of selecting AI + expert 1, through AI + expert $M$); and 3) a \emph{complementary module} \(h_{\psi}: \Delta^{C - 1} \times \mathcal{Y} \to \Delta^{C - 1}\), parametrised by \(\psi \in \Psi\), that aggregates the predictions made by the AI model and a selected human expert to produce a final prediction if the gating model decides to complement AI with one human expert. These components can be visualised in \cref{fig:architecture}.
    
    In standard HAI-CC, ground truth labels are often required for training, while in our setting, ground truth labels are unavailable. 
    Following LECODU~\citep{lecodu}, which also assumes that ground truth labels are unavailable, we use the SOTA MRL method CROWDLAB~\citep{goh2022CROWDLAB} to produce the \emph{consensus} labels to be used as the ground truth in our training.
    CROWDLAB takes the training images and experts' annotations  $(\mathbf{x}_{i}, \mathcal{M}_{i}) \in \mathcal{D}$, together with the AI classifier's predictions $f_{\theta}(\mathbf{x}_{i})$ to produce a consensus label $\hat{y}_{i} \in \mathcal{Y} = \{1, \dots, C\}$ associated with a quality (or confidence) score $\alpha_{i}$.
    Formally, the pseudo clean data set produced by CROWDLAB can be written as follows:
    \begin{equation}
        \begin{aligned}[b]
            & \hat{\mathcal{D}} = \{ (\mathbf{x}_i, \hat{y}_i, \mathcal{M}_i): i \in \{1, \dots, N\} \wedge (\mathbf{x}_i, \mathcal{M}_i) \in \mathcal{D} \wedge \hat{y}_{i} \in \\
            & \{ \hat{y}_{i}: (\hat{y}_{i}, \alpha_i) = \mathsf{CrowdLab}(\mathbf{x}_i, f_{\theta}(\mathbf{x}_i), \mathcal{M}_i) \wedge \alpha_i > 0.5\} \}.
        \end{aligned}
        \label{eq:consensus}
    \end{equation}

Our aim is to learn the parameters of the AI model, the gating model and the complementary module to produce an accurate final prediction, while satisfying the coverage constraint with the following optimisation:
\begin{equation}
    \begin{aligned}[b]
        &\min_{\theta, \phi, \psi} \frac{1}{N}\sum_{i = 1}^{N} g_{\phi}^{\top} (\mathbf{x}_{i}) \pmb{\ell}(\mathbf{x}_{i},  \hat{y}_{i}, \mathcal{M}_{i}, \theta, \psi) \\
        & \qquad \text{s.t.: } \frac{1}{N} \sum_{i = 1}^{N} g^{(\mathrm{AI})}_{\phi}(\mathbf{x}_{i}) \ge \varepsilon,
    \end{aligned}
    \label{eq:loss_function}
\end{equation}
where \(g^{(\mathrm{AI})}_{\phi}(.)\) is the probability of selecting AI alone produced by the gating model, and \(\pmb{\ell}(.)\) is a (\(2M + 1\))-dimensional vector defined as:
\begin{equation}
    \pmb{\ell}(\mathbf{x}_{i},  \hat{y}_{i}, \mathcal{M}_{i}, \theta, \psi) = 
    \begin{bmatrix}
        \ell_{\mathrm{CE}}(\hat{y}_i, f_{\theta}(\mathbf{x}_{i})) \\
        \ell_{\mathrm{CE}}(\hat{y}_i, m_{1}) \\
        \vdots \\
        \ell_{\mathrm{CE}}(\hat{y}_i, m_{M}) \\
        \ell_{\mathrm{CE}}(\hat{y}_i, h_{\psi}(f_{\theta}(\mathbf{x}_i), m_{1}, 1)) \\
        \vdots \\
        \ell_{\mathrm{CE}}(\hat{y}_i, h_{\psi}(f_{\theta}(\mathbf{x}_i), m_{M}, M))
    \end{bmatrix},
    \label{eq:loss_vector}
\end{equation}
with \(\ell_{\mathrm{CE}}\) denoting the cross-entropy loss.

Intuitively, the optimisation in~\cref{eq:loss_function} minimises the weighted-average loss across all available deferral options, with the weights representing the probability produced by the gating model, while the constraint enforces the average probability of selecting the AI classifier alone to be above a certain threshold $\varepsilon$, representing the target coverage. The lower bounded constraint is used in \cref{eq:loss_function} due to the standard assumption in HAI-CC, where human experts generally perform better than the AI classifier. Thus, without imposing such a constraint, the trained gating model will most likely defer to or complement with a human expert without selecting the AI classifier to make the decision alone. In other words, \(g_{\phi}^{(\mathrm{AI})}(\mathbf{x}_{i}) \approx 0\) if there is no constraint.

To optimise the constrained objective in \cref{eq:loss_function}, we use the \emph{penalty method}~\citep[Chapter 17]{nocedal1999numerical}. In particular, the penalty function of the constraint in \cref{eq:loss_function} is defined by
\begin{equation}
    c(\phi, \varepsilon) = \left[ \max \left( 0, \varepsilon - \frac{1}{N} \sum_{i = 1}^{N} g^{(\mathrm{AI})}_{\phi}(\mathbf{x}_{i}) \right) \right]^{2},
    \label{eq:penalty_function}
\end{equation}
which approximates to zero when the constraint is satisfied and becomes positive as the constraint is violated. The loss function in \cref{eq:loss_function} can then be rewritten into a penalty program as follows:
\begin{equation}
    \min_{\theta, \phi, \psi} \frac{1}{N}\sum_{i = 1}^{N} g_{\phi}^{\top} (\mathbf{x}_{i}) \, \pmb{\ell}(\mathbf{x}_{i},  \hat{y}_{i}, \mathcal{M}_{i}, \theta, \psi) + \beta_{k} c(\phi, \varepsilon),
    \label{eq:beta}
\end{equation}
where \(k\) indexes the training iteration, \(\beta_{1} > 0\), and \(\beta_{k + 1} = \lambda (\beta_{k} + k)\),
with \(\lambda > 0\) being a hyper-parameter. Hence, \(\beta_{k+1} > \beta_{k}\) for all training iterations $k$, meaning the training process initially prioritizes minimizing classification loss and gradually shifts its focus toward achieving the target coverage.

The training and testing procedures of CL2DC are summarised in \cref{algorithm:training,algorithm:testing} in the supplementary material. We also depict the inference flow of CL2DC in \cref{fig:architecture}.

\section{Experiments}
\label{sec:experiment}
We evaluate the performance of the proposed method on a variety of datasets including ones with synthetic experts (e.g., CIFAR-100~\citep{krizhevsky2009learning,barz2020we}, HAM10000~\citep{tschandl2018ham10000} and Galaxy Zoo~\citep{bamford2009galaxy}) and real-world ones with labels provided by human experts (e.g., Chaoyang~\citep{zhu2021hard}, MiceBone~\citep{schmarje20192d,schmarje2022one,schmarje2022data} and NIH-ChestXray~\citep{majkowska2020chest,wang2017chestx}).

\begin{figure}[t!]
    \centering
    \includegraphics[width=\linewidth]{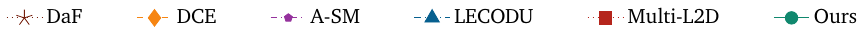}
    \\
    \begin{subfigure}[b]{0.32\linewidth}
        \includegraphics[width=\linewidth]{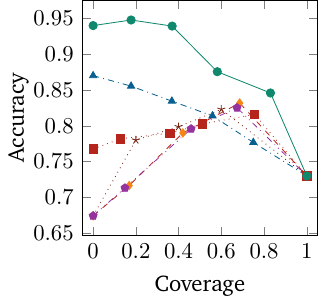}
        \vspace{-1.25em}
        \caption{CIFAR-100}
        \label{fig:cifair100}
    \end{subfigure}
    \hfill
    \begin{subfigure}[b]{0.32\linewidth}
        \centering
        \includegraphics[width=\linewidth]{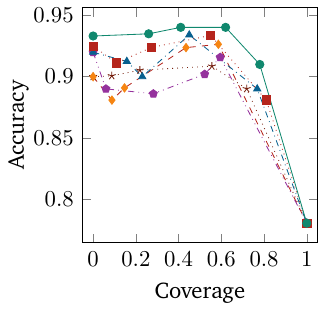}
        \vspace{-1.25em}
        \caption{HAM-10000}
        \label{fig:ham10000}
    \end{subfigure}
    \hfill
    \begin{subfigure}[b]{0.32\linewidth}
        \centering
        \includegraphics[width=\linewidth]{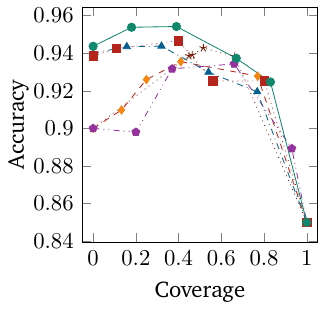}
        \vspace{-1.25em}
        \caption{Galaxy-Zoo}
        \label{fig:galaxyzoo}
    \end{subfigure}
    \\
    \vspace{2em}
    \begin{subfigure}[b]{0.32\linewidth}
        \centering
        \includegraphics[width=\linewidth]{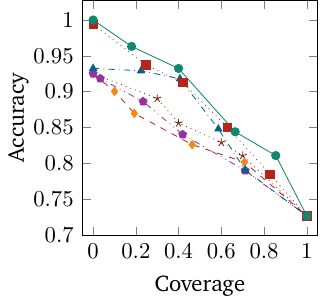}
        \vspace{-1.25em}
        \caption{Chaoyang}
        \label{fig:chaoyang3u}
    \end{subfigure}
    \hfill
    \begin{subfigure}[b]{0.32\linewidth}
        \centering
        \includegraphics[width=\linewidth]{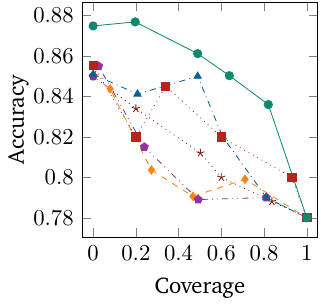}
        \vspace{-1.25em}
        \caption{MiceBone}
        \label{fig:micebone}
    \end{subfigure}
    \hfill
    \begin{subfigure}[b]{0.32\linewidth}
        \centering
        \includegraphics[width=\linewidth]{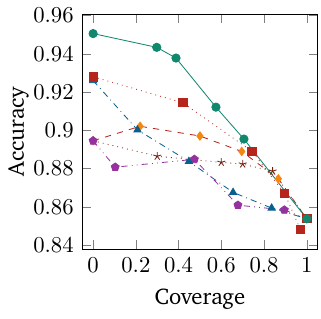}
        \vspace{-1.25em}
        \caption{NIH-AO}
        \label{fig:NIH-AO}
    \end{subfigure}
    \caption{Accuracy-coverage curves of our method and competing SEHAI-CC~\citep{whoshould_mozannar23,charusaiedefer,dce,cao2024defense} and MEHAI-CC~\citep{lecodu,multil2d} methods.}
    \label{fig:exp}
\end{figure}

\subsection{Implementation Details}
\paragraph{Datasets} 

For CIFAR-100, we follow the setting of~\citep{hemmer2023learning} to generate synthetic labels representing  synthetic experts. We generate 3 experts, each one labelling correctly on 6 or 7 different super-classes, while making 50\% labelling mistakes on the remaining 13 or 14 superclasses using asymmetric label noise, where labels can be randomly flipped to other classes within the same super-class. For HAM10000 and Galaxy-zoo, we follow the setting in~\citep{multil2d} to simulate two experts based on two super-classes, each following an asymmetric label noise, similarly to CIFAR-100.

For real-world datasets, we utilise the annotations made by real-world human experts. In Chaoyang dataset, there are 3 human experts with accuracies 91\%, 88\% and 99\%, and we build 2 setups: 1) using the two pathologists with accuracies 88\% and 91\%, forming the setup ``Chaoyang2u''; and 2) using all 3 pathologists, forming the setup ``Chaoyang3u''. In MiceBone dataset, we use 8 out of 79 annotators who label the whole dataset to represent the experts with accuracies varying from 84\% to 86\%. In NIH-ChestXray dataset, each image is annotated by 3 experts who label four radiographic findings. Following~\citep{whoshould_mozannar23,ijcai2022-344}, we focus on the classification of airspace opacity (NIH-AO) because of the balanced prevalence of this finding. The prediction accuracies of the 3 experts in the NIH-AO dataset are approximately 89\%, 94\%, 80\% both in training and testing. Please refer to \cref{sec:supp_datasets,sec:implementation} in the supplementary material for more details on the datasets, architecture and training parameters.

\paragraph{Evaluation}
The evaluation is based on the prediction accuracy as a function of coverage measured on the testing sets. Coverage denotes the percentage of samples classified by the AI model alone, with $100\%$ coverage representing the classification performed exclusively by the AI model, while $0\%$ coverage denoting a classification exclusively done by  experts. All results are computed from the mean result from 3 runs using  the checkpoint obtained at the last training epoch.

\paragraph{Baselines}
We assess our method in both the single and multiple expert human-AI cooperation classification (SEHAI-CC and MEHAI-CC) settings.
For the SEHAI-CC setting, we consider several SOTA methods, such as Asymmetric SoftMax (A-SM)~\citep{cao2024defense}, Dependent Cross-Entropy (DCE)~\citep{dce}, and defer-and-fusion (DaF)~\citep{charusaiedefer} as baselines. For a fair comparison, we randomly sample a single annotation for each image as a way to simulate a single expert from the human annotators to train those SEHAI-CC methods.
For the MEHAI-CC settings, we consider the learning to defer to multiple experts (MultiL2D)~\citep{multil2d} and learning to complement and to defer to multiple experts (LECODU)~\citep{lecodu}.
For a fair comparison, all classification for the \{SE,ME\}HAI-CC methods have the same backbone, and all hyper-parameters are set as previously reported in~\citep{whoshould_mozannar23,lecodu,multil2d}. 
To maintain fairness in the accuracy - coverage comparisons, we include the coverage constraint in \cref{eq:loss_function} into all \{SE,ME\}HAI-CC methods. In particular, we set the hyper-parameter \(\varepsilon\), which controls the coverage lower bound, to \{0, 0.2, 0.4, 0.6, 0.8\} so we can train all methods and plot their corresponding performance using the coverage - accuracy curves.

\subsection{Results}

\begin{table*}[t]
\centering
\resizebox{\linewidth}{!}{
\begin{tabular}{lcccccccc}
\toprule
\bfseries Methods & \bfseries Type & \bfseries CIFAR-100 & \bfseries Chaoyang2u & \bfseries Chaoyang3u & \bfseries NIH-AO & \bfseries Micebone & \bfseries HAM10000 & \bfseries Galaxy-zoo \\
\midrule
A-SM     & SEHAI-CC & 76.41     & 80.84      & 82.02      & 87.33  & 80.31    & 87.51    & 91.28      \\
DCE      & SEHAI-CC & 76.74     & 80.73      & 82.49      & 88.63  & 80.34    & 88.48    & 91.65      \\
DaF      & SEHAI-CC & 77.60     & 81.91      & 84.07      & 88.26  & 81.19    & 88.43    & 91.53      \\
\midrule
MultiL2D & MEHAI-CC & 78.84     & 83.02      & 87.93      & 90.14  & 82.31    & 89.55    & 92.38      \\
LECODU   & MEHAI-CC & 81.17     & 82.89      & 85.71      & 88.19  & 82.47    & 90.15    & 92.67      \\
\rowcolor{gray!25} \bfseries Ours     & MEHAI-CC & \textbf{88.88}     & \textbf{84.79}      & \textbf{88.91}      & \textbf{91.53}  & \textbf{85.27}    & \textbf{91.49} & \textbf{93.57}\\ \bottomrule
\end{tabular}}

\caption{Quantitative comparison in terms of the Area Under Accuracy-Coverage Curve (AUACC)~\citep{nadeem2009accuracy} of the SOTA SEHAI-CC~\citep{whoshould_mozannar23,charusaiedefer,dce,cao2024defense}, MEHAI-CC~\citep{lecodu,multil2d} on the human-AI cooperation classification datasets. The best result per benchmark is marked in bold.}
\label{tab:aucc}
\end{table*}

We report the \textit{accuracy-coverage curves} of the HAI-CC strategies and our proposed method across various datasets in \cref{fig:exp}. These curves illustrate the trade-off between accuracy and cooperation cost as coverage varies from 0\% to 100\%, where 0\% coverage indicates complete reliance on human experts, and 100\% coverage implies classification solely by the AI model. 
Additionally, we provide a concise quantitative analysis of \cref{fig:exp} results in~\cref{tab:aucc}, which shows the \emph{area under the accuracy-coverage curve} (AUACC), where higher AUACC values denote superior accuracy-coverage trade-offs. 

Our method outperforms all competing HAI-CC methods at every coverage level in all benchmarks. Compared with MEHAI-CC methods, the accuracy of SEHAI-CC methods is limited by the lack of specific expert labelling. Consequently, MEHAI-CC methods generally surpass SEHAI-CC approaches, particularly at lower coverage values. However, even in this scenario they still do not match the performance of our method.

\begin{table*}[t!]
\centering
\resizebox{\linewidth}{!}{
\begin{tabular}{l c c c c c c}
\toprule
\multirow{2}{*}{\bfseries Image} & \multicolumn{2}{c}{\bfseries Human prediction} & \bfseries AI prediction  & \bfseries Output of gating model $g_{\phi}(.)$ & \multirow{2}{*}{\bfseries Final prediction} & \multirow{2}{*}{\bfseries GT} \\
\cmidrule{2-3} \cmidrule{5-5}
& \(m_{1}\) & \(m_{2}\) & $f_{\theta}(.)$ & [AI, Ex.1, Ex.2, AI + Ex.1, AI + Ex.2]\({}^{\top}\) & & \\
\midrule
\begin{minipage}[b]{0.1\columnwidth}
	\centering
	\raisebox{-.5\height}{\includegraphics[width=\linewidth]{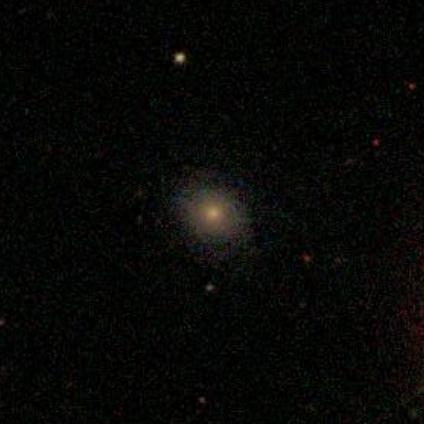}}
\end{minipage} & 1 & 0 & 1     & [0.40, 0.00, 0.01, 0.00, \textbf{0.59}]\({}^{\top}\) & 0         & 0  \\

\begin{minipage}[b]{0.1\columnwidth}
	\centering
	\raisebox{-.5\height}{\includegraphics[width=\linewidth]{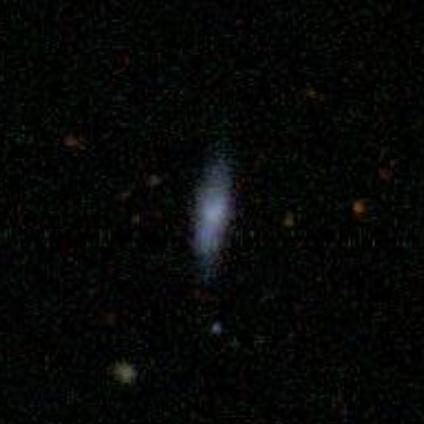}}
\end{minipage} & 0 & 1 & 1     & {[}0.20, 0.00, 0.01, \textbf{0.79}, 0.00{]}\({}^{\top}\) & 0         & 0  \\

\begin{minipage}[b]{0.1\columnwidth}
	\centering
	\raisebox{-.5\height}{\includegraphics[width=\linewidth]{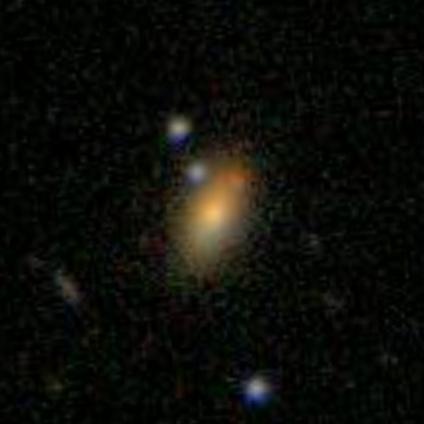}}
\end{minipage} & 1 & 1 & 0     & {[}0.05, 0.00, \textbf{0.94}, 0.00, 0.01{]}\({}^{\top}\) & 1         & 1  \\
\begin{minipage}[b]{0.1\columnwidth}
	\centering
	\raisebox{-.5\height}{\includegraphics[width=\linewidth]{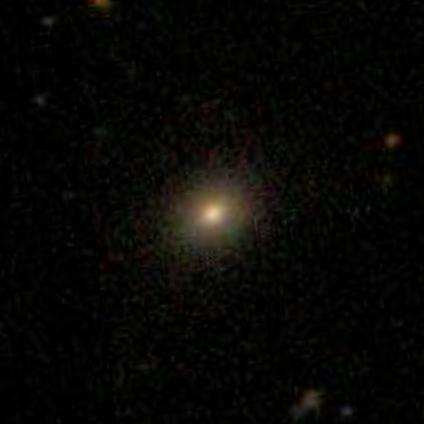}}
\end{minipage} & 1 & 0 & 0     & {[}0.02, \textbf{0.98}, 0.00, 0.00, 0.00{]}\({}^{\top}\) & 1         & 1  \\
\begin{minipage}[b]{0.1\columnwidth}
	\centering
	\raisebox{-.5\height}{\includegraphics[width=\linewidth]{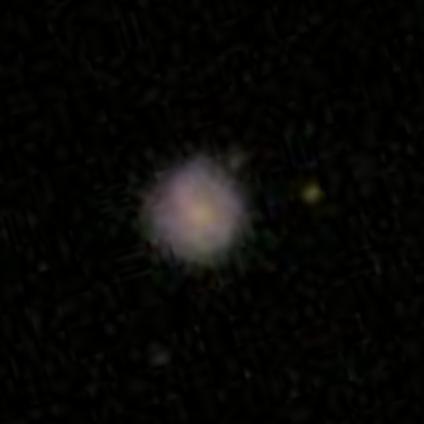}}
\end{minipage} & 0 & 0 & 1     & {[}\textbf{0.80}, 0.00, 0.00, 0.00, 0.20{]}\({}^{\top}\) & 1         & 1  \\
\bottomrule 
\end{tabular}}
\caption{Examples of the CL2DC classification evaluated on the testing samples of the Galaxy Zoo dataset at 40\% coverage. 
}
\label{tab:case}
\end{table*}

In synthetic datasets (i.e., CIFAR-100, Galaxy-zoo and HAM10000), we focus on the setting that different experts have relatively high accuracy on specific categories , as explained in~\cref{sec:supp_datasets} of the supplementary material. CL2DC excels by effectively identifying and cooperating with specific experts for their relevant tasks, thereby optimizing decision-making.
In the CIFAR-100 dataset, LECODU achieves higher accuracy than SEHAI-CC at low and intermediate coverage levels but it shows lower accuracy than our approach. The performance of MultiL2D is comparable to SEHAI-CC, except at 0\% coverage, when MultiL2D is better than SEHAI-CC methods because MultiL2D can successfully identify the best labeller, as opposed to SEHAI-CC methods that pick one of the labellers randomly. 
In the Galaxy-zoo dataset, other MEHAI-CC methods (i.e., LECODU and MultiL2D) show lower accuracy than ours, but they become relatively competitive at higher coverage levels, which underscores our method’s superior adaptability and efficiency in optimising human-AI cooperation. Compared with them, our method excels by effectively identifying and collaborating with specific experts for their relevant tasks, thereby optimising decision-making.

In real-world scenarios (i.e., Chaoyang, Micebone, and NIH-AO), our method consistently outperforms other strategies.
Notably, in Chaoyang, where one of the pathologists has an accuracy close to 100\%, our method adeptly selects this most accurate pathologist, surpassing the performance of LECODU, which randomly selects an expert rather than specifying the optimal one. Although the performance of MultiL2D is competitive in Chaoyang, it is worse than our method in the Micebone and NIH-AO datasets.

\cref{tab:case} shows a few examples of the inference of CL2DC at a coverage rate of 40\% on  test images of Galaxyzoo. Each example includes the test image, the human-provided labels ($\mathcal{M}$), AI model prediction ($f_{\theta}(.)$), complementary module prediction $h_{\psi}(.)$, prediction probability vector by the gating model ($g_{\phi}(.)$), final prediction of CL2DC, and ground truth (GT) label. Notably, when the AI model or the human experts make individual mistakes, the final prediction tends to be correct, highlighting system robustness. When the AI model is correct, $g_{\phi}^{\mathrm{AI}}(.)$ tends to have a high probability suggesting that the AI model can be trusted. 
When the L2D options are selected, that usually happens with very high probability for one of the options in $g_{\phi}^{\mathrm{L2D}_j}(.)|_{j=1}^{M}$ and quite low value for $g_{\phi}^{\mathrm{AI}}(.)$, suggesting a complete lack of trust in the AI model. On the other hand, when one of the L2C options are selected, notice that both $g_{\phi}^{\mathrm{AI}}(.)$ and one of the options in $g_{\phi}^{\mathrm{L2C}_j}(.)|_{j=1}^{M}$ show high values, indicating that the AI model can be partially trusted.

\subsection{Ablation studies}
\label{sec:ablation}
\begin{figure}[tbp]
    \centering
    \subfloat[Ablation study of different \(\lambda\).]{
        \label{fig:ablation_lambda}
        \includegraphics[width=0.45\linewidth]{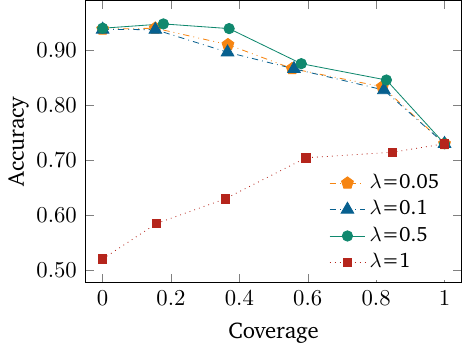}
    }
    \hfill
    \subfloat[Comparison with HAI-CC]{
        \label{fig:chaoyang2u}
        \includegraphics[width=0.45\linewidth]{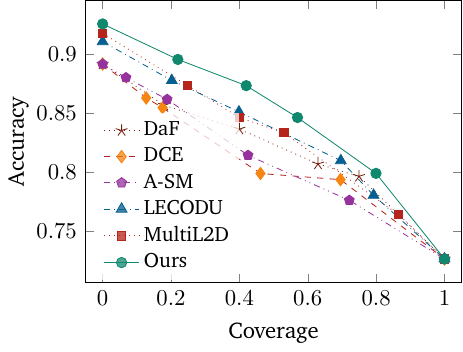}
    }\\
    \subfloat[Varying num. experts on CIFAR-100.]{
        \label{fig:increase_experts_cifair}
        \includegraphics[width=0.45\linewidth]{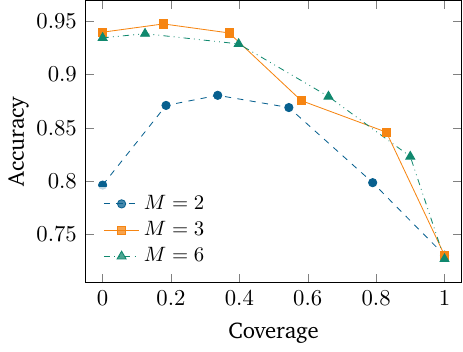}
    }\hfill
    \subfloat[With and w/o L2C on GalaxyZoo.]{
        \label{fig:ablation_l2c_galaxy}
        \includegraphics[width=0.45\linewidth]{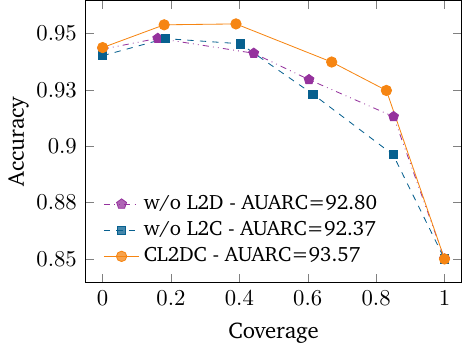}
    }\\
    \subfloat[With and w/o L2C on MiceBone.]{
        \label{fig:ablation_l2c_micebone}
        \includegraphics[width=0.45\linewidth]{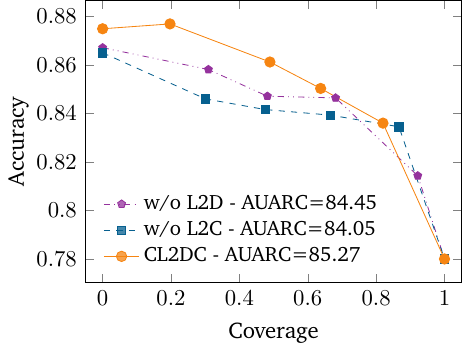}
    }\hfill
    \subfloat[CIFAR-100 with different \textnumero~experts.]{
        \label{fig:area_under_curve_increase_experts}
        \includegraphics[width=0.45\linewidth]{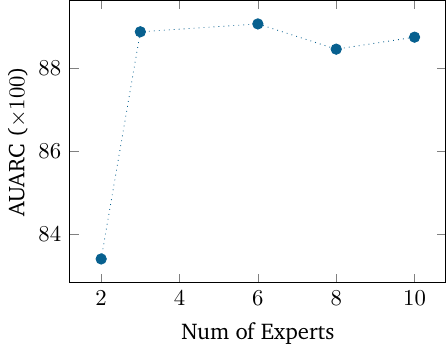}
    }
    \caption{Accuracy-coverage curves in various evaluations: (\subref{fig:ablation_lambda}) different penalty coefficient $\lambda$ on CIFAR-100 dataset, (\subref{fig:chaoyang2u}) comparison between our method and competing HAI-CC methods on Chaoyang dataset with 2 experts, (\subref{fig:increase_experts_cifair}) ablation study when varying the number of experts on CIFAR-100 dataset, (\subref{fig:ablation_l2c_galaxy}) and (\subref{fig:ablation_l2c_micebone}) ablation study with and without L2C and L2C on Galaxyzoo and Micebone, respectively, and (\subref{fig:area_under_curve_increase_experts}) evaluation of AUARC when varying the number of experts on CIFAR-100 dataset.}
\end{figure}

In \cref{fig:ablation_lambda}, we study the hyper-parameter $\lambda$ from~\cref{eq:beta} on the accuracy-coverage performance of CL2DC in  CIFAR-100. The graphs illustrate a clear trend: when $\lambda=1$, the accuracy is distinctively low at lower coverage values because of the relatively high weight for the constraint defined in~\cref{eq:penalty_function}, but when we decrease $\lambda$,
there is an improvement in accuracy for almost all coverage values.

Next, we investigate the effect of altering the number and quality of experts in the experimental setting. Focusing on the ``Chaoyang2u'' setup, the outcomes, displayed in \cref{fig:chaoyang2u} and \cref{tab:aucc}, show that our solution outperforms other HAI-CC methods more distinctly when compared with the original results in \cref{fig:chaoyang3u} that use all three pathologists for the ``Chaoyang3u'' setup.

We then study the influence L2D and L2C in our CL2DC method on Galaxyzoo and Micebone datasets in~\cref{fig:ablation_l2c_galaxy} and~\cref{fig:ablation_l2c_micebone}. 
In~\cref{fig:ablation_l2c_galaxy}, CL2DC w/o L2D outperforms CL2DC w/o L2C at large coverage values, which means that when the expert's accuracy is high, L2C can leverage the accurate expert's prediction, while mitigating the influence of  weak experts by combining their predictions with the AI model prediction. 
CL2DC outperforms CL2DC w/o L2C and CL2DC w/o L2D at all coverage values, showing the advantage of integrating both L2D and L2D into HAI-CC.
In~\cref{fig:ablation_l2c_micebone}, CL2DC w/o L2C performs better than CL2DC w/o L2D, when coverage is larger than 0.6. At a large coverage, L2C may combine a weak expert especially when the expert pool contains a large number of experts who have relatively low accuracies (from 84\% to 86\%). In general, CL2DC tends to work better than CL2DC w/o L2C and CL2DC w/o L2D for most coverage values by leveraging advantages of L2D and L2C.

We also study the scalability of CL2DC when increasing the number of experts on CIFAR-100. 
We generate other seven synthetic experts, who have similar accuracy rates as described in \cref{sec:supp_datasets}, i.e., each expert performs correctly on random 6/7 super-classes, while making 50\% mistakes via instance-dependent noise on the remaining super-classes. 
\cref{fig:increase_experts_cifair} shows the accuracy-coverage curves with our methods, where the number of available experts increases from 2 to 6. In~\cref{fig:area_under_curve_increase_experts}, we evaluate the AUACC as the number of available experts increases. A significant improvement is observed when the number of experts increases from 2 to 3. This is because, in our setting, the first three expert high-accuracy sets of super-classes cover all super-classes, whereas the first two sets cover only about two-thirds of the super-classes. Consequently, it is unsurprising that the AUACC tends to converge as the number of experts continues to increase, indicating that additional redundant experts do not contribute to integrating more effective information or improving predictions.

\section{Conclusion}
\label{sec:conclusion}

In this paper, we propose the novel Coverage-constrained Learning to Defer and Complement with Specific Experts (CL2DC) method. 
CL2DC integrates the strengths of learning-to-defer and learning-to-complement, particularly in training scenarios with multiple noisy-label annotations, enabling the system to either make final decisions autonomously or cooperate with a specific expert. We also introduce and integrate coverage-constraint through an innovative penalty method into the loss function to control the coverage. This penalty allows us to run a robust training procedure where the target coverage can be reached, which consequently enables a reliable analysis of different methods through the coverage - accuracy curves. Comprehensive evaluations across real-world and synthetic multiple noisy label datasets demonstrate CL2DC's superior accuracy to SOTA HAI-CC methods. 

The proposed CL2DC has a limitation with the selection of multiple human experts. In a real-world decision-making system, multiple specific experts may be engaged in the decision process (e.g., clinical diagnosis). In future work, we will develop a new hybrid intelligent system to select a sequence of specific human experts to make decision collaboratively. Another potential issue of CL2DC is the fact that we need roughly balanced training sets to avoid overfitting to majority classes.  We plan to address this problem by leveraging learning methods that are robust to imbalanced distributions.
{
    \small
    \bibliographystyle{ieeenat_fullname}
    \bibliography{main}
}

\newpage
\appendix

\section{Training and testing algorithms}
\label{sec:algorithms}
    We provide the detailed algorithms used in the training and testing of CL2DC in \cref{algorithm:training,algorithm:testing}, respectively.

\begin{algorithm}[tbh]
    \caption{The training procedure of the proposed method}
    \label{algorithm:training}
    \begin{algorithmic}[1]
        \Procedure{CL2DC Training}{$\mathcal{D}, \varepsilon, K, \lambda$}
            \LComment{\(\mathcal{D} = \{(\mathbf{x}_{i}, \mathcal{M}_{i})\}_{i = 1}^{N}\): training dataset consists of samples and annotations of \(M\) human experts}
            \LComment{\(\varepsilon\): the lower bound of the targeted coverage value}
            \LComment{\(K\): the number of iterations}
            \LComment{\(\lambda\): a hyper-parameter used in the penalty function}
            \State initialise the parameters of AI classifier, gating and complement models: \(\theta, \phi\) and \(\psi\)
            \State initialise penalty hyper-parameter: \(\beta_{0} > 0\)
            \State obtain consensus labels: \(( \hat{\mathbf{y}}_{i} )_{i = 1}^{N} \gets \Call{Get consensus labels}{\mathcal{D}, \theta}\)
            \For{\(k = 1:K\)}
                \State initialise \(p_{\mathrm{AI}} \gets 0\) \Comment{to accumulate \say{coverage}}
                \State initialise \(\mathsf{L} \gets 0\) \Comment{to accumulate the training loss}
                \For{each sample indexed by \(i\) in \(\mathcal{D}\)}
                    \State calculate the probability of selection: \(g_{\phi}(\mathbf{x}_{i})\)
                    \State calculate the losses \(\pmb{\ell}(\mathbf{x}_{i}, \hat{\mathbf{y}}_{i}, \mathcal{M}_{i}, \theta, \psi)\) defined in \cref{eq:loss_vector}
                    \State \(\mathsf{L} \gets \mathsf{L} + g_{\phi}(\mathbf{x}_{i})^{\top} \pmb{\ell}(\mathbf{x}_{i}, \hat{\mathbf{y}_{i}}, \mathcal{M}_{i}, \theta, \psi)\) \Comment{accumulate the loss on sample \(i\)-th}
                    \State \(p_{\mathrm{AI}} \gets p_{\mathrm{AI}} + g_{\phi}^{(\mathrm{AI})} (\mathbf{x}_{i})\) \Comment{accumulate coverage}
                \EndFor
                \State calculate the penalty of the constraint: \( c(\phi, \varepsilon) \gets \max \left( 0, \varepsilon - \nicefrac{p_{\mathrm{AI}}}{N} \right)^{2} \) \Comment{defined in \cref{eq:penalty_function}}
                \State \(\beta_{k} \gets \lambda (\beta_{k - 1} + k)\) \Comment{hyper-parameter associated with the penalty function}
                \State \(\mathsf{L} \gets \nicefrac{\mathsf{L}}{N} + \beta_{k} c(\phi, \epsilon)\) \Comment{training loss}
                \State update: \(\theta, \phi, \psi \gets \operatorname{SGD}(\mathsf{L})\)
                
            \EndFor
            \State \Return \(\theta, \phi, \psi\)
        \EndProcedure
        \Statex
        \Procedure{Get consensus labels}{\(\mathcal{D}, \theta\)}
            \LComment{\(\mathcal{D} = \{ (\mathbf{x}_{i}, \mathcal{M}_{i} = \{m_{i}^{(j)}\}_{j = 1}^{M}) \}_{i = 1}^{N}\): the multi-rater noisy label dataset}
            \LComment{\(\theta\): the parameter of a classifier}
            \State get the majority vote label: \(\overline{m}_{i} \gets \Call{majority vote}{\{m_{i}^{(j)}\}_{j = 1}^{M}}, i = 1:\abs{\mathcal{D}} \)
            \State train classifier on majority vote: \(\theta \gets \Call{train classifier}{\{(\mathbf{x}_{i}, \overline{m}_{i})\}_{i = 1}^{\abs{\mathcal{D}}}, \theta}\)
            \State obtain consensus labels: \(( \hat{\mathbf{y}}_{i} )_{i = 1}^{\abs{\mathcal{D}}} \gets \Call{CrowdLab}{\{ (f_{\theta}(\mathbf{x}_i), \mathcal{M}_i) \}_{i = 1}^{\abs{\mathcal{D}}} }\)
            \Comment{defined in \cref{eq:consensus}}
            \State \Return \((\hat{\mathbf{y}}_{i} )_{i = 1}^{\abs{\mathcal{D}}}\)
        \EndProcedure
    \end{algorithmic}
\end{algorithm}

\begin{algorithm}[tbh]
    \caption{The proposed testing procedure}
    \label{algorithm:testing}
    \begin{algorithmic}[1]
        \Procedure{CL2DC Testing}{$(\mathbf{x}_{N + 1}, (m_{N + 1}^{(j)})_{j = 1}^{M}, \theta, \phi, \psi, \varepsilon$}
            \LComment{\(\mathbf{x}_{N + 1}\): a testing sample}
            \LComment{\(m_{N + 1}^{(j)}\): the annotations of the testing samples made by expert indexed by \(j\)}
            \LComment{\(\theta\): parameter of AI classifier}
            \LComment{\(\phi\): parameter of gating model}
            \LComment{\(\psi\): parameter of complementary module}
                \State \( \mathbf{p} \gets g(\mathbf{x}_{N + 1}; \phi) \) \Comment{probability of selection}
                \State \(\mu \gets \operatorname*{argmax}_{\mu} \mathbf{p}_{\mu}\)
                \If{(\(\mu = 1\))} \Comment{AI predict alone}
                    \State \(\tilde{y}_{N + 1} \gets f_{\theta} (\mathbf{x}_{N+1}) \)
                \ElsIf{\(2 \le \mu \le M+1\)} \Comment{learning to defer option}
                    \State \(\tilde{y}_{N + 1} \gets m_{N + 1}^{(\mu - 1)} \)
                \Else \Comment{learning to complement option}
                    \State \(\tilde{y}_{N + 1} \gets h_{\psi}(f_{\theta}(\mathbf{x}_{N + 1}), m_{N + 1}^{(\mu - M - 1)}) \)
                \EndIf
            \State \Return \(\tilde{y}_{N + 1}\)
        \EndProcedure
    \end{algorithmic}
\end{algorithm}

\section{Datasets}
\label{sec:supp_datasets}

\textbf{CIFAR-100}~\citep{krizhevsky2009learning,barz2020we}  has 50k training images and 10k testing images, with each image belonging to one of 100 classes categorised into 20 super-classes. We follow the instance-dependent label noise~\citep{xia2021sample} to generate synthetic labels representing a synthetic expert. In particular, each expert performs correctly on 6 or 7 different super-classes, while making 50\% labelling mistakes on the remaining 13 or 14 super-classes using asymmetric label noise, where labels can be randomly flipped to other classes within the same super-class. In the experiments, we evaluate ours and competing methods using three synthetic experts. In addition, because about 10\% of testing images in CIFAR-100~\citep{krizhevsky2009learning} are duplicated or almost identical to the ones in the training set, in our training and testing, we use ciFAIR-100~\citep{barz2020we}, which replaces those duplicated images by different images belonging to the same class.

\textbf{HAM10000}~\citep{tschandl2018ham10000} has about 10k training and 1,500 testing dermatoscopic images categorised into seven types of human skin lesions. These seven categories can be grouped further into two super-classes: \emph{benign} and \emph{malignant}. We follow the setting presented in Multi-L2D~\citep{multil2d} to simulate two experts based on these two super-classes, each following an asymmetric label noise, similarly to CIFAR-100. 
In particular, the accuracy of the two experts  is around 90\%, where the first expert makes 5\% and 15\% of labelling mistakes on super-classes \emph{malignant} and \emph{benign}, respectively, while the second expert only makes 15\% and 5\% of labelling mistakes on the super-classes \emph{malignant} and \emph{benign}, respectively.

\textbf{Galaxy Zoo}~\citep{bamford2009galaxy} consists of 60k images of galaxies and the corresponding answers from hundreds of thousands of volunteers to classify their shapes.
We follow the setup in Multi-L2D~\citep{multil2d} that uses the response to the first question \emph{"Is the object a smooth galaxy or a galaxy with features/disk?"} as the ground truth labels of a binary classification to simulate two synthetic experts. In particular, the first expert makes 5\% and 15\% of labelling mistakes on \emph{smooth galaxy} and \emph{galaxy with features/disk}, respectively, while the second expert makes 15\% and 5\% of labelling mistakes on \emph{galaxy with features/disk} and \emph{smooth galaxy}, respectively.

\textbf{Chaoyang}~\citep{zhu2021hard} comprises 6,160 colon slide patches categorised into four classes: \emph{normal, serrated, adenocarcinoma, and adenoma}, where each patch has \textit{three noisy labels annotated by three pathologists}. In the original Chaoyang dataset setup, the training set has patches with multi-rater noisy labels, while the testing set only contains patches that all experts agree on a single label. We have restructured the dataset to ensure that 1) \textit{both training and testing sets contain multiple noisy labels}; 2) \textit{experts have similar performance in training and testing sets}; and 3) \textit{patches from the same slide do not appear in both the training and testing sets}. This setting results in a partition of 4,533 patches for training and 1,627 patches for testing, and the accuracy of the three experts are 91\%, 88\%, 99\%, assuming that the majority vote forms the ground truth annotation.

\textbf{MiceBone}~\citep{schmarje20192d,schmarje2022one,schmarje2022data} has 7,240 second-harmonic generation microscopy images, with each image being annotated by one to five professional annotators, where the annotation consists of one of three possible classes: \emph{similar collagen fiber orientation, dissimilar collagen fiber orientation, and not of interest due to noise or background}. Only 8 out of 79 annotators label the whole dataset. We, therefore, use these 8 annotators to represent the experts in our experiment. Using the majority vote as the ground truth, the accuracy of those 8 experts are from 84\% to 86\%. As the dataset is divided into 5 folds, we use the first 4 folds as the training set, and the remaining fold as the test set.

\textbf{NIH-ChestXray}~\citep{majkowska2020chest,wang2017chestx} contains an average of 3 manual labels per image for four radiographic findings on 4,374 chest X-ray images~\citep{majkowska2020chest} from the ChestXray8 dataset~\citep{wang2017chestx}. We focus on the classification of airspace opacity (NIH-AO) because only this finding's prevalence is close to 50\%, without heavy class-imbalance problem. Following~\citep{whoshould_mozannar23,ijcai2022-344}, a total of 2,412 images is for training and 1,962 images are for testing. The prediction accuracy of the 3 experts in the NIH-AO dataset is approximately 89\%, 94\%, 80\% both in training and testing.

\section{Implementation Details}
\label{sec:implementation}
\subsection{Architecture} 

All methods are implemented in PyTorch~\citep{paszke2019pytorch} and run on Nvidia RTX A6000. For experiments performed on CIFAR-100 dataset, we employ ProMix~\citep{wang2022promix} to train the AI model formed by two PreAct-ResNet-18 as the LNL AI models.
For Chaoyang, we use a ResNet-34 for the AI model, and for other datasets, we train the AI model with a ResNet-18 using a regular CE loss minimisation with a ground truth label formed by the majority-voting of experts. The gating model uses the same trained backbones as the ones used for the AI model. The complementary module is represented by a two-layer multi-layer perceptron (MLP), where each hidden layer has 512 nodes activated by Rectified Linear Units (ReLU). On CIFAR-100 the AI model achieves 73.42\% accuracy on the testing set. The AI models on Chaoyang, NIH-AO, Micebone, HAM10000, and Galaxy-zoo datasets achieve 72.65\%, 85.37\%, 78.12\%, 78.06\%, and 85.24\%, respectively. 

\subsection{Training}
For each dataset, the proposed human-AI system is trained for 200 epochs using SGD with a momentum of 0.9 and a weight decay of \(5 \times 10^{-4}\). The batch size used is 256 for all datasets. The initial learning rate is set at 0.01 and decayed through a cosine annealing. 
For training the whole HAI-CC method, the ground truth labels are set as the consensus labels obtained via CROWDLAB. 
For testing, the ground truth label is either available from the dataset (e.g., CIFAR-100, HAM10000, Galaxy-zoo) or from majority voting (e.g., MiceBone, Chaoyang, NIH-ChestXray).

\end{document}